# The Role of Natural Language Understanding in Multimodal Video-Based Dengue Diagnosis

Danial Sharifrazi[1*][0000-0002-8158-0961], Saadat Behzadi[2][0000-0002-8354-0103], Julakha Jahan Jui[1][0009-0001-8476-6788], Mojtaba Mohammadi[1][0000-0002-6443-5930], Nouman Javed[1][0000-0003-0520-3504], Roohallah Alizadehsani[1][0000-0003-0898-5054], Prasad N. Paradkar[3][0000-0002-6553-2214] and Asim Bhatti[1][0000-0001-6876-1437]

[1] Institute for Intelligent Systems Research and Innovations (IISRI), Deakin University, Geelong, Australia
[2] Department of Electronic Engineering, University of Bologna, Bologna, Italy
[3] CSIRO Health and Biosecurity, Australian Animal Health Laboratory, Geelong, Australia
[*] Corresponding author: d.sharifraz@deakin.edu.au

**Abstract.** Detecting infection-related behavioral changes in mosquitoes from video data is challenging because mosquitoes are small, move rapidly and irregularly, and are affected by environmental factors such as background, lighting, and shadows, which can make reliable feature extraction difficult. In this study, a YOLO- and Contrastive Language-Image Pre-training (CLIP)-based vision-language framework is proposed to classify mosquito flight frames of uninfected and Dengue virus serotype 2 (DENV2)-infected mosquitoes. First, YOLO is used to isolate mosquito regions from the background. Then, visual features extracted from video frames are aligned with biologically meaningful textual prompts in a shared embedding space. The multimodal model was fine-tuned using supervised bidirectional contrastive learning and evaluated through frame-level image–text similarity-based classification. The results show that the proposed method achieved 98.54% accuracy and 99.91% sensitivity at the frame level. After temporal aggregation of frame-level information, the model achieved complete video-level performance. The ablation results showed that fine-tuning and CLIP-based representations were essential for this domain, while the textual branch provided semantic image-text alignment rather than an accuracy advantage over the vision-only model. These findings suggest that vision-language models can provide a useful framework for analyzing infection-related biological behaviors from video data.

## 1 Introduction

Analyzing mosquito behavior from video data is a challenging task, especially when the aim is to investigate changes associated with infection. Mosquitoes are very small in most frames and usually occupy only a limited part of the image. Furthermore, their movement is fast, erratic, and sometimes very subtle; therefore, extracting reliable behavioral patterns from such videos is not a simple task. This becomes even more important in distinguishing between infected and uninfected mosquitoes, as mosquitoes are important vectors of arboviruses such as dengue and Zika, and these diseases affect large populations in different regions of the world [1-3].

Biological studies have shown that the effect of viral infections in the vector is not limited to virus replication, but may also affect the nervous system and behavior of the vector. For example, Zika infection has been reported to be associated with neurobehavioral changes in Aedes aegypti [4]. Accordingly, monitoring mosquito flight behavior through video can be a useful approach to study behavioral changes associated with infection. However, factors such as complex backgrounds, lighting changes, shadows, and irregular mosquito movements make it difficult to identify these patterns.

Deep learning methods have been widely used in image, video, and biological data analysis in recent years [5]. In biomedical image analysis, lightweight and real-time frameworks based on YOLO have shown that they can be useful for extracting important visual information from complex images [6]. In addition to image applications, AI-based methods have also been considered in biological and neurophysiological signal analysis, including spiking signal analysis and classification of infection-related neural signals [7, 8]. Also, recent review studies have shown that neurophysiological measurement and modeling play an important role in the development of intelligent and reliable systems [9]. These studies indicate that AI methods can provide effective tools for analyzing complex biological and behavioral patterns. In the mosquito domain, convolutional neural network-based models have been used to count Aedes aegypti eggs and monitor their flight behavior [10, 11]. Also, pre-trained deep models have been used to classify spike sequences associated with dengue and Zika infections in mosquito neurons [8].

To reduce the dependence on extensive labeling, methods such as few-shot learning (FSL), self-supervised learning, contrastive learning, and zero-shot learning have been considered [12-15]. Among these approaches, vision-language models (VLMs) are of particular importance because they can represent images and text in a common semantic space [16]. One of the best-known models in this field is CLIP, which uses contrastive learning to bring related visual and text representations closer together and separate unrelated representations [17]. This feature allows the learning process to be based not only on numerical labels, but also on meaningful text descriptions to guide the model.

Despite the progress of VLMs, their use in mosquito behavior analysis, especially for infection detection from video data, has not yet been widely investigated. Most previous studies have focused on detecting, counting, or tracking mosquito movements [10, 11], while the use of biological textual descriptions to guide model learning has received less attention. On the other hand, methods such as prompt-based learning and adapter-based tuning have shown that VLMs can be adapted for more specialized tasks [18, 19]. Therefore, combining object recognition for mosquito separation from the background with VLMs could provide a suitable framework for analyzing behavioral patterns associated with infection.

Although previous studies have demonstrated the application of Convolutional Neural Network (CNN)-based methods in mosquito monitoring and the application of deep learning models in the analysis of biological signals associated with infection [8, 10, 11], the use of vision-language representations for the analysis of the flight behavior of infected mosquitoes has not yet been widely investigated. This research gap is important because behavioral patterns extracted from video are usually subtle, noisy, and difficult to interpret. In such situations, the use of biological natural-language prompts can provide an additional semantic layer and help improve feature learning and enhance model interpretability.

In this study, a framework based on YOLO and CLIP is presented for classifying flight videos of uninfected and DENV2-infected mosquitoes. First, YOLO is used to identify and separate mosquito-containing regions from the background to reduce the influence of irrelevant environmental factors. Then, the visual representations extracted from the frames are aligned with biologically meaningful text descriptions in a shared space. In the training phase, supervised bidirectional contrastive learning is used to move the samples belonging to each class closer to each other and away from the samples of the other class [20].

The most important achievements of this study are:

- Providing a multimodal framework for classifying infected and uninfected mosquitoes from flight videos;
- Combining YOLO for mosquito-background separation with CLIP for learning VLMs;
- Using biologically meaningful text descriptions to guide the learning process and increase the interpretability of the model;
- Performing frame-level classification through similarity between visual representations and biologically meaningful textual class descriptions.

## 2 Related Works

### 2.1 The Evolution of Deep Learning in Entomological Analytics

In recent years, machine vision and deep learning methods have become increasingly important in entomological studies and vector monitoring. In early studies, deep learning was mostly used for static tasks such as automated mosquito egg counting. For example, EggCountAI showed that CNN-based models can perform accurately even under conditions such as high density and overlapping eggs [10]. However, mosquito analysis is not limited to counting or appearance recognition. Later studies showed that deep models can also be used to track movement and extract subtle behavioral patterns in mosquitoes [8, 11]. This is of particular importance in the context of infection, as biological evidence suggests that viruses such as Zika may affect the nervous system and locomotor behavior of Aedes aegypti [4]. Therefore, video analysis of mosquito flight behavior could be a promising avenue for investigating behavioral changes associated with infection.

### 2.2 Limitations of Labeled Data and the Move to Prompt-Guided Methods

Despite the progress of deep learning methods, one of their main limitations is their heavy dependence on labeled data. In biological and entomological studies, the preparation of labeled data is usually time-consuming, expensive, and dependent on expert knowledge. This problem is exacerbated in video data, as there may be a need to examine frame-by-frame or analyze movement patterns.

To reduce this dependence, methods such as zero-shot learning, contrastive learning, and prompt-based models have been considered [15, 20]. In these approaches, the model uses semantic and textual descriptions to understand visual features instead of learning only numerical labels. This feature can be useful in mosquito behavior analysis, as biological information can be fed into the learning process in the form of textual descriptions.

### 2.3 VLMs and the gap in mosquito behavior analysis

VLMs such as CLIP have enabled the representation of images and text in a shared embedding space [16, 17]. In these models, related visual and textual representations are brought closer together, and unrelated examples are spaced apart. This structure allows the model to use natural-language prompts to guide the recognition process and, in some situations, to perform reasonably well without the need for extensive labeled data.
In recent years, methods such as CoOp and CLIP-Adapter have shown that VLMs can be tuned for more specialized tasks using prompt learning or lightweight adaptive layers [18, 19]. Also, more recent studies in other fields have shown that textual descriptions can guide the model's attention to more important features [21-23].

Despite these advances, the use of VLMs in analyzing mosquito behavior, especially for detecting infection from video, is still limited. Most previous studies have focused on counting, detecting, or tracking mosquitoes, while combining biological knowledge in the form of textual prompts with visual features extracted from video has been less explored. To address this gap, this study presents a framework based on YOLO and CLIP, in which mosquito-background separation is combined with vision-language learning and biological textual prompts.

## 3 Dataset

The data used in this study were obtained by continuously monitoring 15 mosquitoes in a laboratory chamber. In order to provide nutrients and ensure the survival of the mosquitoes during the imaging period, a food source containing a water and sugar solution was placed inside the chamber. To record behavioral patterns, a camera was placed in front of the cage and data recording was continued for 1-13 days under different lighting conditions.

This dataset contains a total of 60 videos recorded from the flight paths of mosquitoes in the confined environment of the cage. Each video contained the entire group of 15 mosquitoes rather than an individually recorded mosquito. In contrast to the large volume of raw data, these videos are carefully classified into two main classes: uninfected mosquitoes (control group) and mosquitoes infected with dengue fever virus. To maintain balance in training the model, the number of samples in both classes is considered to be completely equal (30 videos for each group).

In these images, there are several environmental challenges that make feature extraction difficult; These include light fluctuations in the laboratory environment, uneven light distribution in the left and right parts of the cage, and slight changes in camera angles. In addition, factors such as the subjects being too small and their irregular and chaotic movements in the frame complicate the extraction of biological patterns. These factors may lead to the creation of an incorrect embedding space in the model, which in turn challenges the validity of the classification process and negatively affects the final results. The main goal in this section is to prepare this data to separate the morphological and behavioral characteristics of mosquitoes from the environmental noise so that the model can distinguish between the two healthy and infected classes with high accuracy.

## 4 Proposed Methodology

In this section, a proposed VLM framework for binary classification of mosquito flight behavior is presented. The goal is to distinguish between uninfected or control samples and DENV2-infected samples. The proposed method consists of several main steps: data preprocessing using object detection, using the CLIP base model with the clip_vit_base_patch32 architecture, designing a supervised two-way contrastive loss function, and evaluating the model at the frame level using cross-validation. The overall workflow of the proposed framework is illustrated in Fig. 1.

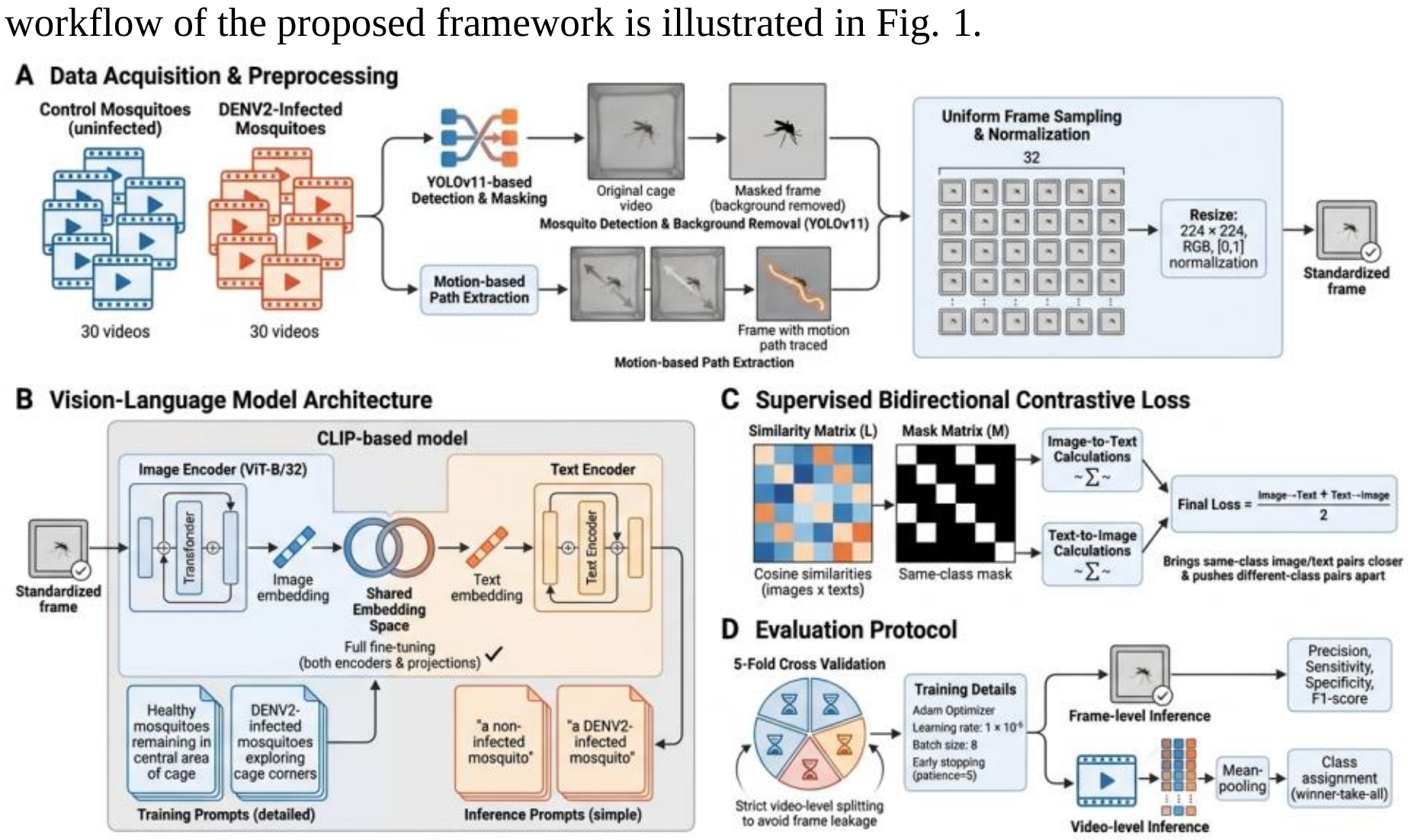

Fig. 1: Overview of the proposed framework, including data preprocessing, CLIP-based image-text alignment, supervised bidirectional contrastive learning, and frame-level evaluation.

**A. Object and Motion Detection**

The dataset consists of 60 videos of mosquito flight paths in a controlled cage environment. These videos are balanced between two groups of control mosquitoes and DENV2-infected mosquitoes. A spatial preprocessing step was performed to reduce the effect of environmental noise and focus the model on the appearance and behavioral characteristics of the mosquito.

In the main image processing scenario, the YOLOv11 model was used to detect and track mosquitoes in video frames. Then, a masking method was used to remove the background, leaving only the mosquito region in the image. This reduced the effects of factors such as cage structure, light, shadow, and other irrelevant components. In the second scenario, instead of YOLO-based masking, a motion-based method was used that indirectly extracts the mosquito's movement path based on the difference between consecutive frames.

To standardize the model input and reduce the computational cost, a fixed number of frames was selected from each video. Specifically, T = 32 frames were uniformly sampled from each video. If the number of frames in a video was less than T, all available frames were used; however, for longer videos, frames were uniformly selected from the video length. Finally, the frames were resized to 224 × 224, converted to RGB color space, and normalized to the interval [0, 1].

**B. Proposed Vision-Language Architecture**

To map the flight visual patterns and biological natural-language prompts into a shared embedding space, the CLIP model with the clip_vit_base_patch32 architecture was used. The model consists of a Vision Transformer-based image encoder and a text encoder that embeds the image and text representations into a common multi-faceted space.

To select the best configuration, four different strategies were investigated. In the first strategy, both image and text encoders, along with the projection layers, were fully fine-tuned. In the second strategy, both encoders were kept fixed and only the projection layers were trained. In the third strategy, the text encoder was kept fixed, but the image encoder and the projections were trained. In the fourth strategy, in addition to the full fine-tuning of the encoders, an LSTM layer was added to the output of the image encoder to model the temporal dependencies between consecutive frames.

Based on the experimental results, the first strategy showed the best performance and representation stability. In this configuration, the full fine-tuning of the image encoder allows the model to learn specific features related to the flight behavior of mosquitoes. Also, training the text encoder makes the semantic space of the text compatible with the biological descriptions used in this study. For this reason, this configuration was chosen as the final architecture of the proposed method.

**C. Semantic Prompt Generation and Textual Embedding**

In VLMs, the text input plays the role of a semantic guide and helps the model organize visual representations based on the biological concept of each class. In this study, the prompts were designed to reflect both the infection status and the behavioral pattern observed in the data.

In the training phase, more descriptive prompts were used, such as: "Healthy mosquitoes remaining in the central area of the cage" and "DENV2-infected mosquitoes exploring cage corners". These prompts were biologically motivated by previously reported arbovirus-associated behavioral changes in Aedes aegypti [4] and were used as hypothesis-driven semantic descriptions.

These prompts combine the biological state of each group with its associated locomotor or spatial pattern. In the inference phase, simpler and more direct prompts were used: "a non-infected mosquito" for the control class" and "a DENV2-infected mosquito" for the DENV2 class.

All text sequences were processed using the CLIP model tokenizer and the Byte Pair Encoding method. To equalize the length of the inputs, the tokens were padded or truncated to a maximum of 20 tokens. These tokens were then converted into dense vectors by the text encoder and, after projection, were placed in a shared embedding space. Finally, the text representations were prepared for calculating similarity with the image representations using L2 normalization.

**D. Supervised Bidirectional Contrastive Loss**

In conventional contrastive learning, it is usually assumed that each image is paired with only one corresponding text in the batch. However, in this problem, several different frames may belong to the same class and have the same prompt. Therefore, if the conventional loss is used, the model may mistakenly consider some examples of the same class as negative examples. This can cause a false negative penalty.

To solve this problem, a supervised bidirectional contrastive loss function was used in this study. First, a normalized mask matrix M was constructed, which indicates whether image *i* and text *j* belong to the same class or not. If *y* is the vector of actual labels for the batch of size B, this mask is defined and normalized as follows:

$$M_{i,j} = \frac{\mathbb{I}(y_i = y_j)}{\sum_{k=1}^{B} \mathbb{I}(y_i = y_k)}$$

where $\mathbb{I}$ denotes the indicator function.

Then, a similarity matrix L of size B × B was calculated based on the cosine similarity between the image and text embeddings. These embeddings were standardized with L2 normalization before calculating the similarity. Using the normalized mask M as the target distribution, cross-entropy was calculated in two directions: from image to text and from text to image. The image-to-text loss and text-to-image loss were defined as follows:

$$\mathcal{L}_{I \to T} = -\frac{1}{B}\sum_{i=1}^{B}\sum_{j=1}^{B} M_{i,j} \log \left(\frac{\exp(L_{i,j})}{\sum_{k=1}^{B} \exp(L_{i,k})}\right)$$

$$\mathcal{L}_{T \to I} = -\frac{1}{B}\sum_{i=1}^{B}\sum_{j=1}^{B} M_{i,j} \log \left(\frac{\exp(L_{j,i})}{\sum_{k=1}^{B} \exp(L_{k,i})}\right)$$

Therefore, the final objective function was defined as the arithmetic mean of the two two-way losses:

$$\mathcal{L}_{total} = \frac{1}{2}(\mathcal{L}_{I \to T} + \mathcal{L}_{T \to I})$$

This formulation causes the same-class samples in the embedding space to be closer together and the samples belonging to different classes to be further apart. Therefore, the model can better distinguish the flight behavior of control and DENV2-infected mosquitoes.

**E. Experimental Setup and Evaluation Protocol**

To evaluate the generalizability of the model and reduce the probability of overfitting, a 5-fold cross-validation was used. Data splitting was performed at the video level so that frames of the same video were not included in the training and test sets at the same time. At each fold, a part of the training data was also separated for validation.

The model was trained with the Adam optimizer. The batch size was set to 8 and the learning rate was set to $1 \times 10^{-5}$. Also, to prevent overfitting, early stopping with a patience of 5 epochs was used based on the training loss.

Performance was evaluated at the frame level:

**Frame-Level Inference:** Individual frames were projected into the latent space and classified based on the highest cosine similarity to the inference text prompts. Metrics including Precision, Sensitivity, Specificity, and F1-score were computed.

# 5 Results

In this section, the performance of the proposed framework for detecting the flight behavior of DENV2-infected mosquitoes is evaluated. To examine the stability and generalizability of the model, all experiments were conducted using 5-fold cross-validation. The results are reported at the frame level across five cross-validation folds, and the proposed method is compared with two alternative strategies, including a motion-based method and an LSTM-based model.

## 5.1 Performance Analysis of the Proposed Method

Table 1 summarizes the overall performance of the proposed model at the frame level. As can be seen, the model achieved an accuracy of 98.54%, a sensitivity of 99.91%, and an F1-score of 98.28% at the frame level. The high value of sensitivity indicates that the proposed framework was able to identify infected samples at the frame level with high accuracy. Also, the appropriate value of specificity and precision indicate that the model, in addition to detecting infected samples, also had an acceptable performance in reducing errors related to control samples.

Table 1. Overall frame-level performance of the proposed model

| Accuracy | Sensitivity | Specificity | Precision | F1 Score |
|---|---|---|---|---|
| 98.54% | 99.91% | 97.55% | 96.75% | 98.28% |

To examine the robustness of the model across different data partitions, the fold-wise results are reported in Table 2. This table shows that the model performance was high and stable across most folds. Although some folds, such as Fold 5, show a relative decrease in specificity and precision, the sensitivity of the model remained at a very high level across all folds. This indicates that the model is reliable in identifying the DENV2-infected class, even when there are differences in image quality, motion complexity, or video recording conditions.

Table 2. Fold-wise frame-level performance of the proposed model

| Metrics / Fold | Accuracy | Sensitivity | Specificity | Precision | F1-score |
|---|---|---|---|---|---|
| 1 | 100% | 100% | 100% | 100% | 100% |
| 2 | 98.7% | 100% | 97.4% | 97.5% | 98.7% |
| 3 | 98.18% | 100% | 97.3% | 94.8% | 97.3% |
| 4 | 99.5% | 99.6% | 99.4% | 99.6% | 99.6% |
| 5 | 96.4% | 100% | 93.8% | 92% | 95.8% |

### 5.2 Training Dynamics

Fig. 2 shows the training process of the model over 5 folds. The accuracy and loss curves indicate the rapid and relatively stable convergence of the model. The rapid increase in contrastive accuracy in the early epochs, along with the gradual decrease in loss, indicates that the model is able to align the visual and text representations in the shared space well.

Also, the closeness of the training and validation trends indicates that the model does not show severe overfitting. However, due to the limited size of the dataset, this result should be interpreted with caution and evaluation on larger data is necessary to verify the stability of the model.

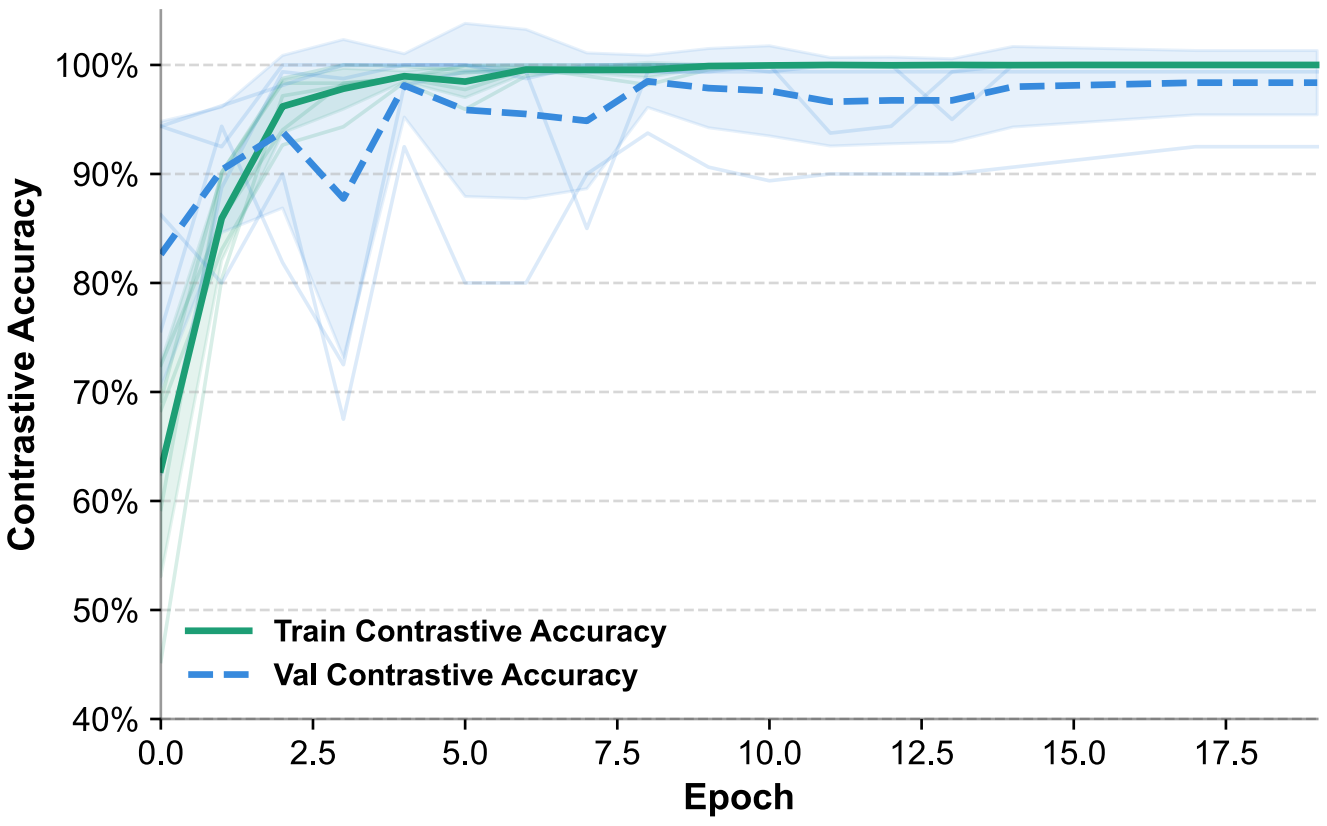

Fig. 2. Training dynamics of the proposed model across 5-fold cross-validation

### 5.3 Comparison with Motion-Driven and LSTM-Based Methods

To investigate the effectiveness of the proposed method, its performance was compared with two alternative methods, including motion-driven and LSTM-based models. The results of this comparison are reported in Table 3 for the frame level.

At the frame level, the motion-driven method achieved strong specificity and precision, whereas the LSTM-based configuration showed very low sensitivity and F1-score under the current training setting. The proposed method achieved the highest sensitivity and F1-score while maintaining high specificity and precision, resulting in the most balanced performance profile among the evaluated configurations. Fig. 3 summarizes these frame-level results.

Table 3. Comparison of motion-driven, LSTM-based, and proposed methods.

| Training Strategy | Accuracy | Sensitivity | Specificity | Precision | F1 Score |
|---|---|---|---|---|---|
| Motion driven | 97.19% | 94.97% | 99.56% | 99.74% | 97.21% |
| LSTM-based | 43.59% | 0.07% | 80.60% | 0.16% | 0.10% |
| **Proposed** | **98.54%** | **99.91%** | **97.55%** | **96.75%** | **98.28%** |

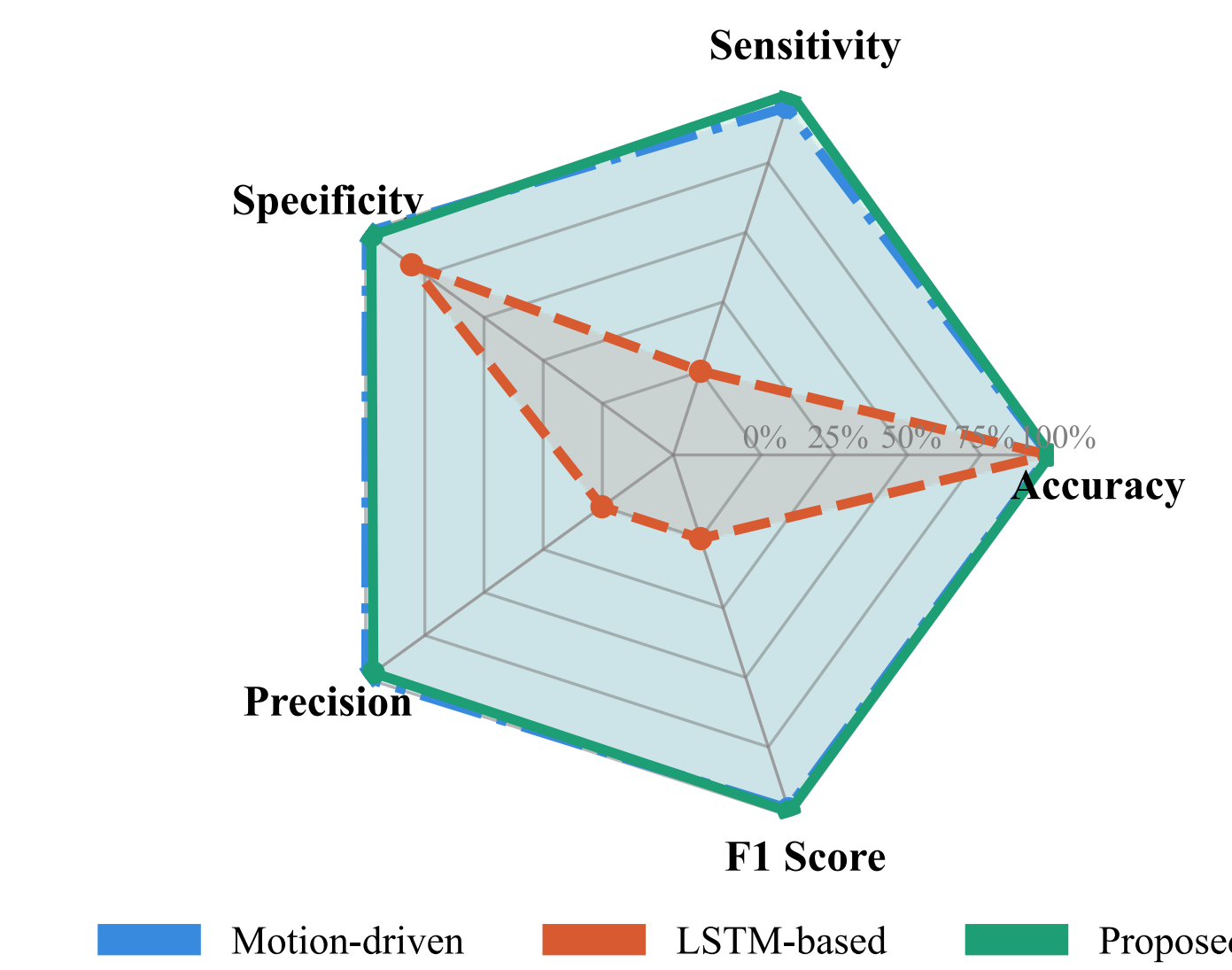

Fig. 3. Radar plot comparison of motion-driven, LSTM-based, and proposed methods.

### 5.4 Ablation Study

In this section, several ablation experiments were conducted to investigate the role of the main components of the proposed framework. The aim of these analyses is to determine the impact of fine-tuning, image-text alignment, and the use of a pre-trained CLIP image encoder on the final performance of the model.

**Analysis of the training strategy: frozen weights versus full fine-tuning.** In the first experiment, the effect of updating the CLIP weights was investigated. For this purpose, the model performance was evaluated in the case where both the image and text encoders were completely frozen and the model used only the prior knowledge of CLIP. As shown in Table 4, this strategy caused the model to completely fail to identify DENV2-infected mosquitoes. Although the accuracy at the frame level showed a seemingly acceptable value, the sensitivity and F1-score were equal to zero. This result indicates that the model assigned almost all samples to the control class.

This finding suggests that the general features of CLIP are not sufficient for this biological problem. The flight behavior and subtle changes caused by infection in mosquitoes are probably not well represented in the general CLIP pre-training data. Therefore, full fine-tuning is necessary to adapt the model to this domain. The confusion matrix shows that the frozen model is unable to correctly identify the DENV2-infected class and attributes most of the samples to the control class. As shown in Fig. 4, the frozen CLIP model assigns all samples to the control class, whereas the proposed model clearly distinguishes between the two classes.

**Analysis of the role of multimodal alignment.** In the second experiment, the role of aligning visual features with text descriptions was investigated. To do this, the performance of the proposed model was compared with an image-only model that uses only visual features and does not use text prompts.

Table 4. Effect of freezing CLIP encoder weights on model performance.

| Configuration model | Accuracy | Sensitivity | Specificity | Precision | F1 Score |
|---|---|---|---|---|---|
| Frozen CLIP | 50.0% | 0% | 100% | 0% | 0% |

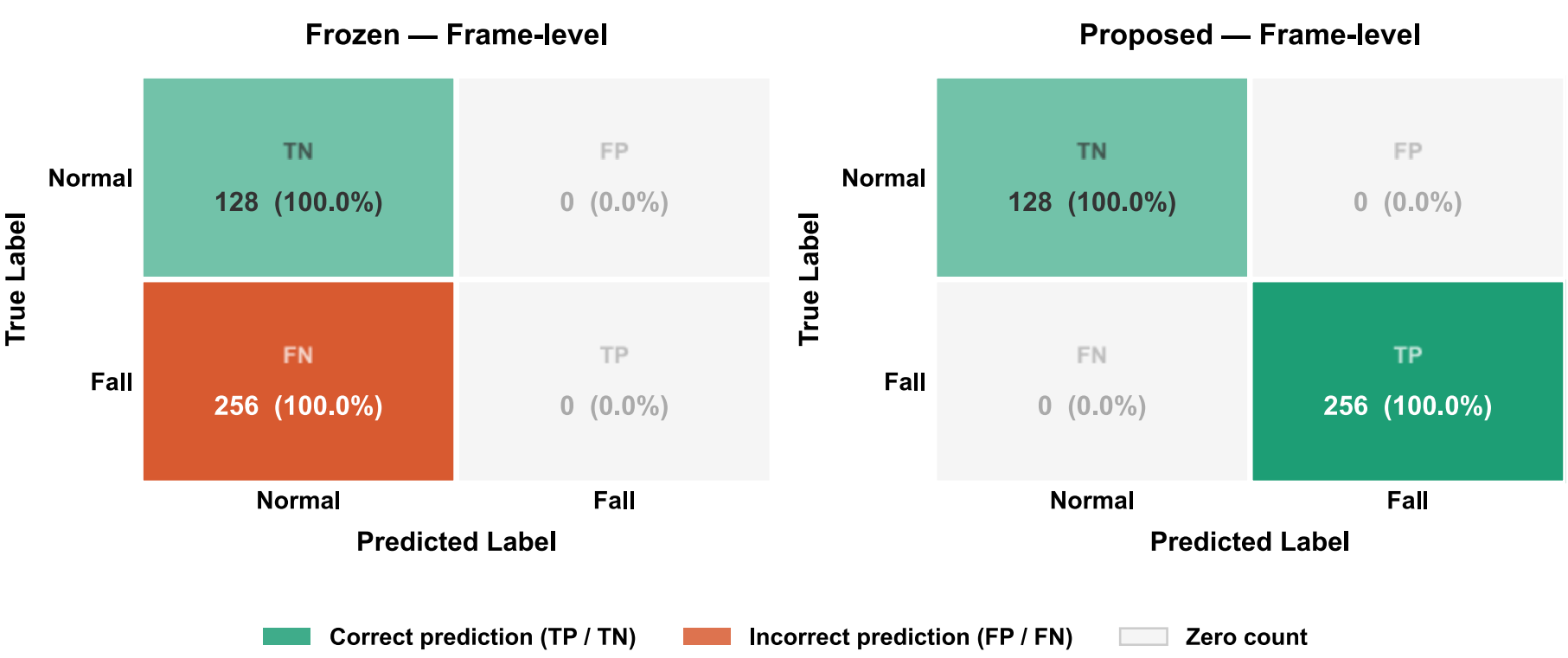

Fig. 4. Confusion matrices of frozen CLIP and proposed models

As shown in Table 5, the image-only model achieved frame-level performance comparable to the proposed model. Therefore, the textual component should not be interpreted as providing a clear accuracy advantage; its primary role is to align visual representations with semantically meaningful textual descriptions.

**Investigating the effect of the pre-trained visual encoder.** In the third experiment, the effect of the pre-trained visual encoder CLIP was investigated. For this purpose, the vision-only model, which uses only the ViT-B/32 visual encoder, was compared with the image-only model. The results in Table 5 show that the vision-only configuration achieves strong frame-level performance.

Table 5. Frame-level performance comparison of Image-only and Vision-only configurations.

| Configuration model | Accuracy | Sensitivity | Specificity | Precision | F1 Score |
|---|---|---|---|---|---|
| Image-only | 97.60% | 100% | 96.22% | 94.38% | 96.94% |
| Vision-only | 99.01% | 99.80% | 98.56% | 97.80% | 98.76% |

These results show that the pre-trained CLIP visual encoder provides strong representations for detecting subtle mosquito flight patterns. The comparable performance of the vision-only and multimodal configurations indicates that the textual branch does not provide a measurable accuracy advantage in the current dataset. Instead, it enables classification through semantic image–text alignment rather than a conventional fixed classification head.

Fig. 5 compares the frame-level performance of the Image-only, Vision-only, and Proposed configurations. The trainable configurations achieve comparable results, while the proposed model additionally represents class decisions through image–text similarity.

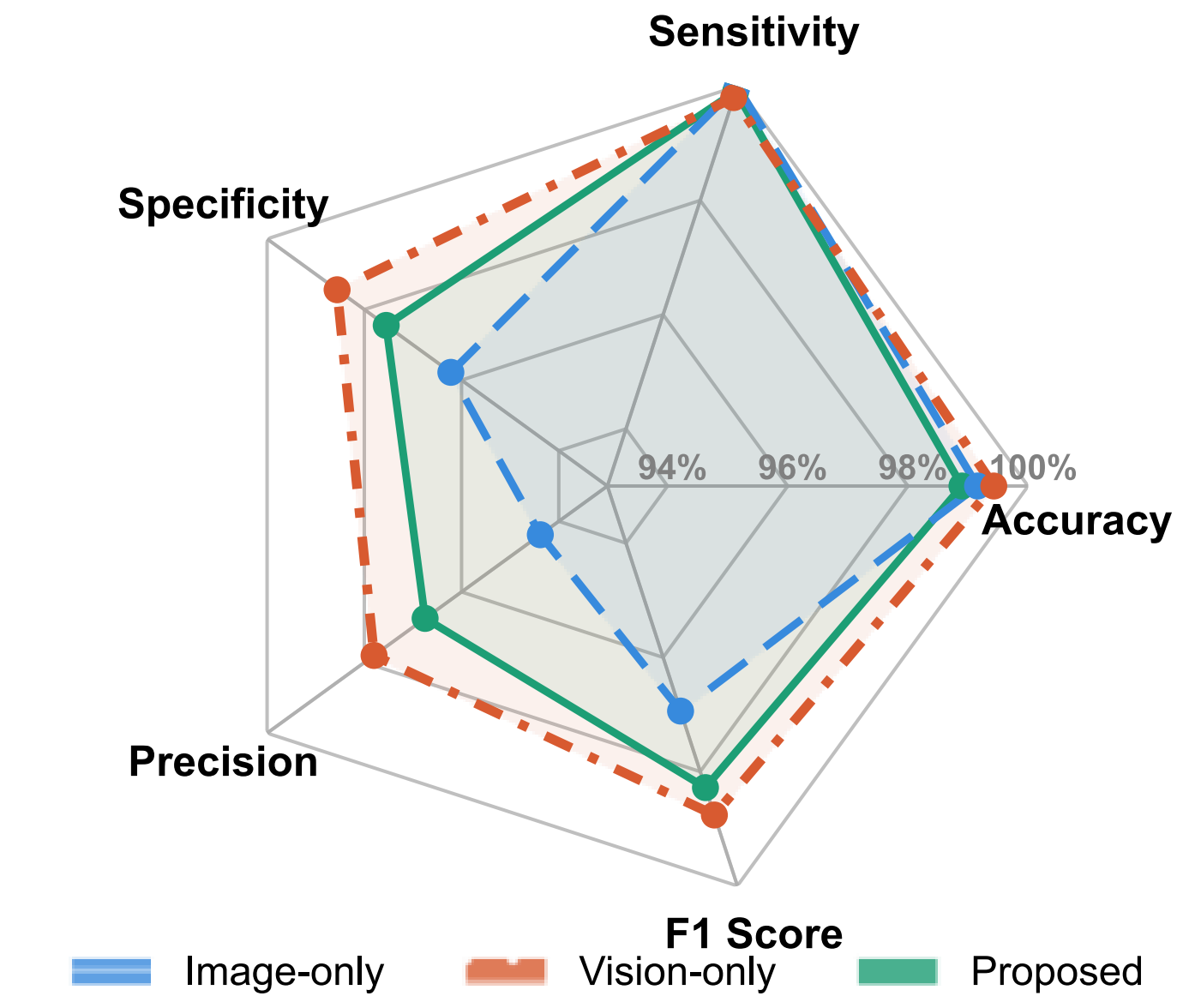

Fig. 5. Radar plot comparison of image-only, vision-only, and proposed models.

## 6 Discussion

The proposed method showed strong performance at the frame level across the five-fold cross-validation. The sensitivity of the model remained very high across all folds, although a relative decrease in specificity and precision was observed in some folds, particularly Fold 5. These results indicate that the model is capable of extracting visual patterns associated with the control and DENV2-infected groups under the investigated imaging conditions. However, variations in image quality, mosquito movement complexity, lighting, and recording conditions may affect the separation of the two classes.

Compared with the alternative methods, the motion detection-based method remained competitive at the frame level, whereas the LSTM-based model showed very low sensitivity and F1-score under the current setting. The proposed method achieved higher sensitivity and F1-score while maintaining high specificity and precision, resulting in the most balanced performance among the evaluated methods. However, these comparisons should be interpreted as controlled internal comparisons conducted using the same dataset and evaluation protocol, rather than as a comprehensive comparison with all existing methods.

The ablation results also clarified the roles of the different model components. Keeping both CLIP encoders completely frozen caused the model to fail to identify the infected class. This finding suggests that general CLIP representations alone are insufficient for this biological problem and that domain-specific fine-tuning is necessary. On the other hand, the Vision-only and multimodal models achieved comparable classification performance, with the Vision-only model performing slightly better in some metrics. Therefore, the textual component should not be interpreted as directly improving classification accuracy. Its main role is to create a shared image–text space in which model decisions are associated with meaningful biological descriptions, rather than being generated only by a fixed visual classification head. This structure provides a more interpretable and semantic interface for representing class concepts, although its ability to identify new concepts by simply changing the prompt was not evaluated in this study.

The main limitations of this study are the limited number of recordings and the limited biological diversity of the dataset. Each video contained the entire group of mosquitoes and was not associated with a single individual mosquito. To prevent direct information leakage, all frames extracted from the same video were assigned to the same cross-validation fold. However, frames within a video remain temporally and visually correlated and should not be considered completely independent biological observations. Therefore, the present results should be interpreted as preliminary evidence under the evaluated conditions. Future evaluation using a larger number of recordings, larger and more diverse mosquito cohorts, different imaging conditions, and alternative prompts could provide a more reliable assessment of the method's generalizability.

## 7 Conclusion

In this study, a multimodal framework based on YOLO and CLIP was proposed for frame-level classification of flight patterns associated with control and DENV2-infected mosquitoes. First, YOLO was used to isolate mosquito-related regions and reduce the influence of the background. The fine-tuned CLIP model then aligned the visual representations with biologically meaningful textual descriptions in a shared embedding space. The proposed method showed strong and balanced frame-level performance across five-fold cross-validation.

The ablation results showed that pretrained CLIP features without domain-specific fine-tuning were insufficient for this task and that fine-tuning was necessary to adapt the model to the biological domain. In addition, the comparable performance of the Vision-only and multimodal models indicated that the textual component did not provide a distinct accuracy advantage. Instead, its main contribution was to provide a shared semantic representation between images and text and a decision-making mechanism based on textual descriptions.
Despite the promising results, the limited number of recordings and the correlations between frames extracted from the same recording remain the main limitations of this study. Further evaluation on larger datasets, more diverse imaging conditions, and larger and more biologically diverse mosquito cohorts is needed to provide a more reliable assessment of the method's generalizability.

**Disclosure of Interests.** The authors have no competing interests to declare that are relevant to the content of this article.